\documentclass[runningheads]{llncs}

\usepackage{cite}
\usepackage[T1]{fontenc}
\usepackage{graphicx}
\usepackage{amsmath}
\usepackage{multirow}
\usepackage[normalem]{ulem}
\useunder{\uline}{\ul}{}
\usepackage{booktabs}
\usepackage{xspace}

\newcommand{\modelname}{MStar\xspace}
\DeclareMathOperator*{\argmax}{arg\,max}

\usepackage{dsfont}
\usepackage{bbm}
\usepackage{enumitem}
\usepackage{amsfonts,amssymb}
\usepackage{orcidlink}

\usepackage[misc]{ifsym}

\begin{document}
\title{Expanding the Scope:\\ Inductive Knowledge Graph Reasoning with Multi-Starting Progressive Propagation}
\titlerunning{Inductive Knowledge Graph Reasoning with MStar}
\author{Zhoutian Shao\inst{1}\orcidlink{0009-0003-6277-268X} \and
Yuanning Cui\inst{1}\orcidlink{0000-0002-9113-0155} \and
Wei Hu\inst{1,2}$^{(\textrm{\Letter})}$\orcidlink{0000-0003-3635-6335}
}
\authorrunning{Z. Shao et al.}
\institute{
State Key Laboratory for Novel Software Technology,\\
Nanjing University, Nanjing 210023, China 
\and
National Institute of Healthcare Data Science,\\
Nanjing University, Nanjing 210093, China\\
\email{ztshao.nju@gmail.com, yncui.nju@gmail.com, whu@nju.edu.cn}\\
}

\maketitle

\begin{abstract}
Knowledge graphs (KGs) are widely acknowledged as incomplete, and new entities are constantly emerging in the real world.
Inductive KG reasoning aims to predict missing facts for these new entities. 
Among existing models, graph neural networks (GNNs) based ones have shown promising performance for this task.
However, they are still challenged by inefficient message propagation due to the distance and scalability issues.
In this paper, we propose a new inductive KG reasoning model, \textbf{\modelname}, by leveraging conditional message passing neural networks (C-MPNNs).
Our key insight is to select multiple query-specific starting entities to expand the scope of progressive propagation.
To propagate query-related messages to a farther area within limited steps, we subsequently design a highway layer to propagate information toward these selected starting entities.
Moreover, we introduce a training strategy called LinkVerify to mitigate the impact of noisy training samples.
Experimental results validate that \modelname achieves superior performance compared with state-of-the-art models, especially for distant entities.
\end{abstract}

\section{Introduction}
Knowledge graphs (KGs) have become a valuable asset for many downstream AI applications, including semantic search, question answering, and logic reasoning~\cite{downstream, reviewer2_survey1, reviewer2_survey2}.
Real-world KGs, such as Freebase~\cite{freebase}, NELL~\cite{nell}, and DBpedia~\cite{dbpedia}, often suffer from the incompleteness issue that lacks massive certain triplets~\cite{new_add_ref1, new_add_ref2}.
The KG reasoning task aims to alleviate incompleteness by discovering missing triplets based on the knowledge learned from known facts.
Early studies \cite{trans_survey} assume that KGs are static, ignoring the potential unseen entities and emerging triplets in the continuously updated real-world KGs.
This motivates the task of inductive KG reasoning \cite{grail,downstream3}, which allows for incorporating emerging entities and facts during inference.

Due to their excellent efficiency and performance, conditional message passing neural networks (C-MPNNs), such as NBFNet~\cite{NBFNet} and RED-GNN~\cite{RED_GNN}, have emerged as one of the premier models in the field of inductive KG reasoning.
To transmit conditions, existing C-MPNNs only incorporate conditional information into the head entity and propagate along the relational paths progressively.
However, this single-starting-entity strategy results in a limited conditional message passing scope, leading to the failure of message passing from the head entity to distant target entities.
This inspires us to extend the scope of conditional message passing to support reasoning on target entities in a farther area.

We conduct an empirical study to analyze the drawbacks of the limited message passing scope. 
Specifically, we report the results of a C-MPNN, RED-GNN~\cite{RED_GNN}, on predicting target entities at different distances in Table~\ref{table:sample}.
It can be observed that RED-GNN performs poorly for queries with distant target entities, even stacking more message-passing layers.
This indicates that existing C-MPNNs cannot effectively propagate conditional messages toward distant target entities, hindering performance on these queries.
Although stacking more GNN layers can alleviate this issue to some extent, it causes high computation and performance declines on the queries with target entities nearby.

\begin{table}[!t]
\centering
\caption{
Hits@10 results of RED-GNN~\cite{RED_GNN} in our empirical study. 
We divide all triplets in the FB15k-237 (v1) dataset~\cite{grail} into four groups according to the shortest distance between head and tail entities. 
``$\infty$'' denotes that the head entity cannot reach the corresponding tail entity in the KG. 
The maximum shortest distance is 10 in the FB15k-237~(v1) when ignoring triplets belonging to $\infty$.
}
\label{table:sample}
\setlength{\tabcolsep}{4pt}
\begin{tabular}{ccccc}
\toprule
Distance & Proportions & Layers\,$=$\,3 & Layers\,$=$\,6 & Layers\,$=$\,9 \\
\midrule
$[1,\ 4)$ & 70.25\% & .611 & .594 & .587 \\
$[4,\ 7)$ & 22.44\% & .000 & .102 & .154 \\
$[7, 10]$ & \ \,3.90\% & .000 & .000 & .088 \\
$\infty$  & \ \,3.41\% & .000 & .000 & .000
\\
\bottomrule
\end{tabular}
\end{table}

\begin{figure}[!t]
\centering
\includegraphics[width=\textwidth]{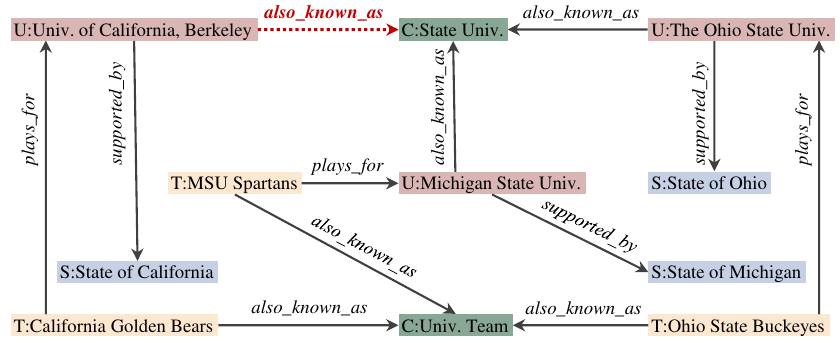}
\caption{
A motivating example of distant target tail entities for predicting (\textit{Univ. of California, Berkeley} $\rightarrow$ \textit{also\_known\_as} $\rightarrow$ \textit{State Univ.}).
Prefix ``U'', ``S'', and ``T'' represent university, state, and basketball teams, respectively.
Prefix ``C'' represents category-type entities.
Different colors and prefixes symbolize distinct entity types.
}
\label{fig:sample}
\end{figure}

In this paper, we propose a novel inductive KG reasoning model \modelname based on \textbf{M}ulti-\textbf{Star}ting progressive propagation, 
which expands the scope of efficient conditional message passing.
Our key insight is to utilize more conditional starting entities and create shortcuts between the head entity and them.
Specifically, we design a starting entities selection (SES) module and a highway layer to select multiple starting entities and create shortcuts for conditional message passing, respectively.
First, the SES module encodes entities using a pre-embedded GNN and then selects multiple query-dependent starting entities, which may include entities distant from the head entity.
These entities broaden the scope of subsequent progressive propagation and allow \modelname to propagate along query-related relational paths to enhance reasoning concerning distant entities.
Second, we create shortcuts from the head entity to the selected multiple starting entities in the highway layer.
The design of the highway layer is inspired by skip connection from ResNet~\cite{resnet}.
The conditional message can be passed to distant entities through the highway layer.
For example, in Fig.~\ref{fig:sample}, a 3-layer RED-GNN would fail to predict the target answer, because the length of the shortest path between head entity \textit{Univ. of California, Berkeley} and the target entity \textit{State Univ.} is larger than 3.
In contrast, our \modelname can select multiple starting entities, e.g., \textit{Michigan State Univ.} and \textit{The Ohio State Univ.}, and transmit conditional messages to them through the highway layer. 
Thus, \modelname can achieve a better reasoning performance than other C-MPNNs on this query.
After the highway layer, we follow it with a multi-condition GNN to perform message passing based on the embeddings of multiple starting entities.
We also propose a training sample filtering strategy called LinkVerify to reduce the impact of the unvisited target entities.
Overall, \modelname visits more query-related distant entities in limited steps and provides more conditional information to these entities compared with existing models.

Our main contributions in this paper are summarized as follows:
\begin{itemize}
\item 
We propose a novel inductive KG reasoning framework based on C-MPNN, named \textbf{\modelname}.
It extends the scope of conditional message passing to improve the predictions of distant target entities.

\item  
We design two modules, SES and highway layer.
The SES module performs starting entities selection for visiting distant entities.
The highway layer provides shortcuts for efficient conditional message passing, alleviating computation waste during additional propagation.

\item 
We conduct extensive experiments on inductive datasets to demonstrate the effectiveness of our framework and each module.
The results show that \modelname outperforms the existing state-of-the-art reasoning models and improves the performance on queries with distant target entities.
\end{itemize}

The rest of this paper is organized as follows.
We first discuss related works in Section~\ref{section:related_work}.
Then, we describe the reasoning task and propagation mechanisms in Section~\ref{section:preliminaries}.
The details of \modelname are presented in Section~\ref{section:methodology}, and the experimental results are reported in Section~\ref{section:experiments}.
Finally, in Section~\ref{section:conclusion}, we discuss the superiority of \modelname and possible extensions in future work.

\section{Related Work}
\label{section:related_work}
\subsection{Knowledge Graph Reasoning}
KG reasoning has been an active research area due to the incompleteness of KGs.
Typical KG reasoning models process each triplet independently and extract the latent semantics of entities and relations.
To model the semantics of the triplets, TransE~\cite{transe}, TransH~\cite{transh}, TransR~\cite{transr}, and RotatE~\cite{rotate} compute translational distance variously.
RESCAL~\cite{rescal}, DistMult~\cite{distmult}, and ComplEx~\cite{complex} follow another reasoning paradigm based on semantic matching.
Instead of exploring the information implied in a single triplet, R-GCN~\cite{rgcn} and CompGCN~\cite{compgcn} capture global structure evidence based on graph neural networks (GNNs).
These models, however, learn unary fixed embedding from training, which cannot be generalized to emerging entities in the inductive KGs.
Instead, our model embodies relational information to encode emerging entities.

\subsection{Inductive Knowledge Graph Reasoning}
One research line of inductive KG reasoning is rule mining, independent of entity identities.
RuleN~\cite{rulen} and AnyBURL~\cite{anyburl} try to prune the process of rule searching.
Neural LP~\cite{neural_lp} and DRUM~\cite{drum} propose to learn logical rules in an end-to-end differentiable manner, learning weights for each relation type and path.
However, the rules are usually short due to the expensive computation for mining and may not be generalized to distant entities.

Another research line is subgraph extraction.
GraIL~\cite{grail} extracts subgraphs around each candidate triplet and labels the entities with the distance to the head and tail entities.
CoMPILE~\cite{compile}, TACT~\cite{tact}, SNRI~\cite{snri}, LogCo~\cite{logco}, and ConGLR~\cite{conglr} follow a similar subgraph-labeling paradigm.
However, the subgraphs that these models extract convey insufficient information due to sparsity.
These models constitute our baselines for inductive KG reasoning.

\subsection{Conditional Message Passing Neural Networks}
Recently, a variant of GNNs called conditional message passing neural networks (C-MPNNs)~\cite{CMPNN} propagates messages along the relational paths and encodes pairwise entity embeddings.
Given a query head $u$ and a query relation $q$ as conditions, C-MPNNs compute embeddings of $(v\,|\,u,q)$ for all entity $v$.
To incorporate conditions into embeddings, NBFNet~\cite{NBFNet} and A*Net~\cite{A_Net} initialize the head entity with the embedding of query relation and propagate in the full KG for each GNN layer.
However, conditional information passing is still restricted in the neighborhood of the head entity.
Differently, RED-GNN~\cite{RED_GNN}, AdaProp~\cite{AdaProp}, and RUN-GNN~\cite{RUN_GNN} propagate the message progressively starting from the head entity without special initialization.
During progressive propagation, the involved entity set is augmented step by step with the neighbor entities of the current set instead of being a full entity set.
Thus, progressive propagation cannot even visit distant entities in limited steps.
\modelname alleviates the above problem by selecting multiple starting entities adaptively for progressive propagation and transmitting conditional information through shortcuts.

EL-GNN~\cite{el_gnn} is another work related to C-MPNNs.
This study proposes that C-MPNNs learn the rules of treating the head entity as constant when the head entity is initialized with conditional information.
Thus, EL-GNN learns more rules by assigning unique embeddings for entities whose out-degree in the KG reaches a specific threshold.
However, the degree and entity-specific embeddings are fixed, which violates the nature of inductive KG reasoning.
Our \modelname selects starting entities according to the query and generates conditional entity embeddings, which can be applied to unseen entities.

\subsection{Skip Connection}
Skip connection~\cite{resnet} is a popular technique in deep learning that skips one or more layers.
Skipping layers contributes to addressing vanishing or exploding gradients~\cite{skip_connection2} by providing a highway for the gradients.
ResNet~\cite{resnet} constructs the highway by adding input $\mathbf{x}$ and output $F(\mathbf{x})$.
DenseNet~\cite{densenet} provides multiple highways by concatenating the input of each layer.
These models transmit the input in shallow layers directly to the target deeper layer in an efficient way.
Inspired by skip connection, \modelname constructs a highway with several new edges to transmit messages faster and propagate to farther entities.

\section{Preliminaries}
\label{section:preliminaries}
\subsubsection{Knowledge Graph}
A KG $\mathcal{G}=(\mathcal{E}, \mathcal{R}, \mathcal{F})$ is composed of finite sets of entities $\mathcal{E}$, relations $\mathcal{R}$, and triplets $\mathcal{F}$.
Each triplet $f\in\mathcal{F}$ describes a fact from head entity to tail entity with a specific relation, i.e., $f=(u, q, v) \in \mathcal{E} \times \mathcal{R} \times \mathcal{E}$, where $u$, $q$, and $v$ denote the head entity, relation, and tail entity, respectively.

\subsubsection{(Inductive) Knowledge Graph Reasoning}
To complete the missing triplet in real-world KGs, KG reasoning is proposed to predict the target tail entity or head entity with a given query $(u, q, ?)$ or $(?, q, v)$.
Given a source KG $\mathcal{G}=(\mathcal{E}, \mathcal{R}, \mathcal{F})$,
inductive KG reasoning aims to predict the triplets involved in the target KG $\mathcal{G}^{\prime}=(\mathcal{E}^{\prime}, \mathcal{R}^{\prime}, \mathcal{F}^{\prime})$, where $\mathcal{R}^{\prime}\subseteq \mathcal{R}$, $\mathcal{E}^{\prime} \not\subset \mathcal{E}$, and $\mathcal{F}^{\prime} \not\subset \mathcal{F}$.

\subsubsection{Starting Entities in Progressive Propagation}
GNNs transmit messages based on the message propagation framework~\cite{propagation1, propagation2}.
This framework prepares an entity set to transmit messages for each propagation step.
Full propagation transmits messages among all entities at all times.
Progressive propagation continuously incorporates the neighbor entities of the entity set in the previous step.
Based on progressive propagation, we use starting entities to indicate the entities involved in the first layer of the GNN.
Given the starting entities $\mathcal{S}$, the entities involved in the $\ell^{th}$ layer of the GNN can be formulated as
\begin{equation}
\mathcal{V}^{\ell} = \left\{
\begin{array}{lc} 
\mathcal{S} & \quad\ell =0 \\
\mathcal{V}^{\ell-1} \cup 
\left\{x\,|\,\exists (e, r, x) \in \mathcal{N}(e) \wedge e \in \mathcal{V}^{\ell-1}\right\}
& \quad\ell > 0
\end{array}
\right.,
\nonumber
\end{equation}
where $\mathcal{N}(e)$ denotes the neighbor edges of the entity $e$.
In particular, NBFNet puts all the entities into $\mathcal{S}$, i.e., $\mathcal{S}=\mathcal{E}$.
RED-GNN only puts the head entity into $\mathcal{S}$, i.e., $\mathcal{S}=\{u\}$ with given query $(u,q,?)$.
Too few starting entities limit the scope of conditional message passing.
On the contrary, too many start entities disperse the attention of GNNs on local information which is critical for reasoning.
Our model \modelname strikes a balance by including the head entity and some selected query-dependent starting entities that are helpful for reasoning.

\section{Methodology}
\label{section:methodology}

\begin{figure}[!t]
\centering
\includegraphics[width=\textwidth]{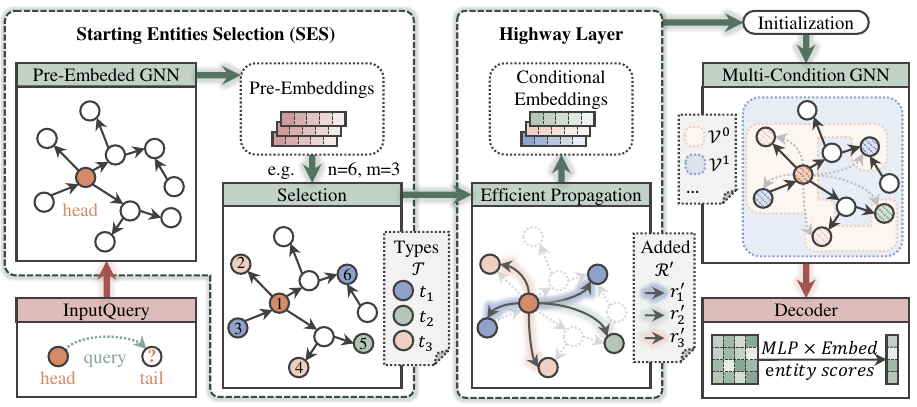}
\caption{Framework overview of \modelname} \label{fig:framework}
\end{figure}

\subsection{Model Architecture Overview}
The overview of \modelname is presented in Fig.~\ref{fig:framework}.
Specifically, we first employ the pre-embedded GNN to pre-encode all entities.
Then, SES selects $n$ query-dependent starting entities according to the pre-embeddings.
The highway layer classifies starting entities into $m$ types, considering the correlation between the head entity and other starting entities.
To improve message-passing efficiency, the highway layer maps each entity type into a new relation and constructs shortcut edges between the head entity and other starting entities.
Based on the message passing on the shortcut edges, we use the highway layer to obtain conditional entity embeddings as the initialization for multi-condition GNN.
Finally, the multi-condition GNN propagates relational information progressively conditioned on these starting entities and generates pairwise embeddings of each entity.
According to the final entity embeddings, the decoder operates as a multilayer perceptron (MLP) and generates scores for each candidate entity.

\subsection{Starting Entities Selection}
As shown in Fig.~\ref{fig:sample}, progressive propagation starts from the only entity (head entity) and cannot reach the distant entities.
However, the excessive utilization of starting entities introduces noisy relational paths into the reasoning.
Despite the expansion of the propagation, some starting entities still miss the target entities and visit other distant entities unrelated to the query.
Thus, we propose to select multiple query-dependent starting entities adaptively to cover a farther area but not introduce irrelevant noise in reasoning.

\subsubsection{Pre-Embedded GNN}
To find the starting entities related to the query, we first introduce a pre-embedded GNN to learn the simple semantics of the entities.
The pre-embedded GNN transmits messages among all entities in the KG following the full propagation paradigm.
To explore query-related knowledge, the pre-embedded GNN encodes the relation conditioned on query relation $q$.
Specifically, the computation for message passing is given by 
\begin{align}
\mathbf{h}_{\text{pre}|u,q}^{\ell}(e)&=\dfrac{1}{|\mathcal{N}(e)|}\sum_{(e,r,x)\in \mathcal{N}(e)}
\Big(\mathbf{h}_{\text{pre}|u,q}^{\ell-1}(x)+\hat{\mathbf{r}}_q\Big),\\
\hat{\mathbf{r}}_q&=\mathbf{W}_{r}\mathbf{q}+\mathbf{b}_{r},
\label{equ:relative_r}
\end{align}
where 
$\mathbf{h}_{\text{pre}|u,q}^{\ell}(e)$ denotes the embedding of the entity $e$ in propagation step $\ell$, 
$\mathbf{q}$ is a learnable embeddings for relation $q$, 
$\mathbf{W}_{r}\in \mathbb{R}^{d \times d}$ is an $r$-specific learnable weight matrix, 
and $\mathbf{b}_{r} \in \mathbb{R}^{d}$ is an $r$-specific learnable bias.
$d$ is the dimension of both entity and relation embeddings.
$\hat{\mathbf{r}}_q$ denotes the embedding of relation $r$ conditioned on $q$.
The pre-embedded GNN initializes $\mathbf{h}_{\text{pre}|u,q}^{0}$ as zero vectors and produces the entity embeddings $\mathbf{h}_{\text{pre}|u,q}^{L_{1}}$ after $L_1$ layers of message passing.

\subsubsection{Selection}
Provided with the embeddings of entities conditioned on $u$ and $q$, we design a score function to select query-dependent starting entities.
The score function measures the importance of entities relative to the head entity and query relation.
Given an entity $e$, the importance score $\alpha_{e|u,q}$ is defined as
\begin{equation}
    \alpha_{e|u,q} = \mathbf{W}_1\left(\text{ReLU}\left(
    \mathbf{W}_2\left(
        \mathbf{h}_{\text{pre}|u,q}^{L_1}(e) \oplus 
        \mathbf{h}_{\text{pre}|u,q}^{L_1}(u) \oplus 
        \mathbf{q}
    \right)
\right)\right),
\end{equation}
where $\mathbf{W}_1 \in \mathbb{R}^{1\times d}$ and $\mathbf{W}_2 \in \mathbb{R}^{d\times 3d}$ are learnable weight matrices.
$\oplus$ denotes the concatenation of two vectors.
We keep the top-$n$ entities as starting entity set $\mathcal{S}_{u,q}$.
$\mathcal{S}_{u,q}$ can propagate along the relational paths conditioned on the query.

\subsection{Highway Layer}
Given multiple starting entities, progressive propagation can traverse more entities, particularly those located at distant positions.
The distant entities, however, receive nothing about the conditional information, due to the limited scope of conditional message passing.
Inspired by the skip connection~\cite{resnet}, which allows skip-layer feature propagation, we introduce a highway layer to tackle this issue.

Aiming to propagate conditional information to the starting entities, we consider constructing shortcut edges between the query head entity and the other starting ones.
Due to the different semantics of the starting entities, we classify entities into $m$ types based on the embeddings.
Each type indicates that this group of entities has a specific semantic relationship with the head entity.
Then, we map each entity type to a new semantic relation type and construct new edges.
Given conditions $u$, $q$ and entity $e$, the entity type is defined as follows:
\begin{equation}
    \beta_{e|u,q} = \argmax_t \mathbf{W}_t \mathbf{h}_{\text{pre}|u,q}^{L_1}(e), \quad t \in [1, m],
\end{equation}
where $t$ is a type of starting entities, and $\mathbf{W}_t \in \mathbb{R}^{1\times d}$ is a $t$-specific learnable weight matrix.

Given starting entity types, the highway layer constructs shortcut edges as
\begin{equation}
    \mathcal{H}_{u,q} = \left\{(u,r_{\beta_{e|u,q}}^{\prime},e)\,|\,e \in \mathcal{S}_{u,q} - \{u\} \right\},
\end{equation}
where $r_{\beta_{e|u,q}}^{\prime}$ denotes the new relation that we introduce, corresponding to the starting entity type.
These edges act as a skip connection to support skipping propagation from the head to the starting entities.

Finally, the highway layer performs message passing on $\mathcal{H}_{u,q}$ to obtain the embeddings of the selected starting entities:
\begin{equation}
    \mathbf{g}_{u,q}(e) = \sum_{(e,r,x)\in \mathcal{N}_{\text{highway}}(e)}
    \mathbf{g}_{u,q}(x)\odot\hat{\mathbf{r}}_q,
\end{equation}
where $\mathbf{g}_{u,q}(e)$ denotes the embedding of entity $e$, 
$\mathcal{N}_{\text{highway}}(e)$ denotes the neighbor edges of the entity $e$ in set $\mathcal{H}_{u,q}$,
and $\odot$ denotes the point-wise product between two vectors.
To satisfy target entity distinguishability~\cite{CMPNN}, we set a learnable embedding for the head entity $u$.

\subsection{Multi-Condition GNN}
In \modelname, we introduce a multi-condition GNN to produce the final entity embeddings.
The multi-condition GNN is a C-MPNN conditioned on the head entity and query relation.
Specifically, the multi-condition GNN initializes entity embeddings $\mathbf{h}_{u,q}^{0}$ as $\mathbf{g}_{u,q}$ and propagates from the starting entities progressively.
Given the query triplet $(u,q,?)$, we incorporate the query information into propagation in two ways.

First, we model the embedding of relation $r$ in an edge as $\hat{\mathbf{r}}_q$ conditioned on the query relation $q$ same as Eq.~(\ref{equ:relative_r}).
Second, considering that the semantics of edges are query-dependent, we use the attention mechanism~\cite{attn} and assign a weight for every edge $(e,r,x)$ in step $\ell$:
\begin{equation}
\gamma^{\ell}_{(e,r,x)|u,q} = \sigma\Big(
\mathbf{W}_{\text{attn}}^{\ell}\text{ReLU}
    \big(
        \mathbf{W}^{\ell}_{\text{attn}_{\text{u}}} \mathbf{h}_{u,q}^{\ell-1}(e) +
        \mathbf{W}^{\ell}_{\text{attn}_{\text{r}}} \hat{\mathbf{r}}_q +
        \mathbf{W}^{\ell}_{\text{attn}_{\text{q}}} \mathbf{q}
    \big)
\Big),
\end{equation}
where 
$\mathbf{W}^{\ell}_{\text{attn}} \in \mathbb{R}^{1\times d_{\gamma}}$, $\mathbf{W}^{\ell}_{\text{attn}_{\text{u}}}$, 
$\mathbf{W}^{\ell}_{\text{attn}_\text{r}}$ and
$\mathbf{W}^{\ell}_{\text{attn}_\text{q}} \in \mathbb{R}^{d_{\gamma} \times d}$ are learnable weight matrices, 
$d_{\gamma}$ is the dimension of attention,
$\mathbf{h}_{u,q}^{\ell}(e)$ denotes the embedding of the entity $e$ in multi-condition GNN at step $\ell$,
and $\sigma$ denotes a sigmoid function.

Based on the two ways above, the entity embeddings are given by
\begin{equation}
\mathbf{h}_{u,q}^{\ell}(e) = \text{ReLU}\left(
\mathbf{W}_o^{\ell} 
\sum_{(e,r,x) \in \mathcal{N}(e) \wedge
\{e, x\}\subset \mathcal{V}^{\ell}_{u,q}
}
\gamma^{\ell}_{(e,r,x)|u,q} 
\left(
    {\mathbf{h}}_{u,q}^{\ell-1}(x) \odot \hat{\mathbf{r}}_q
\right)
\right),
\end{equation}
where $\mathbf{W}_o^{\ell} \in \mathbb{R}^{d\times d}$ is a learnable weight matrix,
$\mathcal{V}^{\ell}_{u,q}$ is the entity set in progressive propagation step $\ell$, and $\mathcal{V}^{0}_{u,q} = \mathcal{S}_{u,q}$.

\subsection{Training Strategy: LinkVerify}
To reason the likelihood of a triplet $(u,q,e)$, the decoder produces a score function $\operatorname{s}(\cdot)$.
Given the final output ${\mathbf{h}}_{u,q}^{{L}_2}$ after $L_2$ layers of multi-condition GNN, the score function is given by
\begin{equation}
\operatorname{s}\left(u,q,e\right)= \mathbf{W}_3
\left(\text{ReLU}
\left(\mathbf{W}_4
    \left(
        {\mathbf{h}}_{u,q}^{L_2}(u)
        \oplus 
        {\mathbf{h}}_{u,q}^{L_2}(e)
    \right)
\right)
\right),
\end{equation}
where 
$\mathbf{W}_3 \in \mathbb{R}^{1\times d}$ and $\mathbf{W}_4 \in \mathbb{R}^{d\times 2d}$ are learnable weight matrices.
However, multi-condition GNN propagates progressively and probably misses several distant target tail entities during the training.
In this situation, the prediction knows nothing about the target tail entity and brings a noisy score for training.

To alleviate the problem above, we propose a mechanism \textbf{LinkVerify} to filter noisy training samples.
The noisy sample represents the triplet whose target tail entity is not involved in $\mathcal{V}^{L_2}_{u,q}$.
Taking the inductive KG reasoning task as a multi-label classification problem, we use the multi-class log-loss~\cite{loss, RED_GNN} to optimize the model.
Associated with LinkVerify, the final loss is given by
\begin{equation}
\mathcal{L}=\sum_{(u,q,v)\in\mathcal{F}}
\bigg(
    -\operatorname{s}\left(u,q,v\right)+
    \text{log}\Big(
        \sum_{\forall e\in\mathcal{E}} \exp\big(\operatorname{s}(u,q,e)\big)
    \Big)
\bigg)
\times\mathds{1}\left(v \in \mathcal{V}^{L_2}_{u,q}\right).
\end{equation}

\section{Experiments}
\label{section:experiments}
In this section, we perform extensive experiments to answer the questions below:
\begin{itemize}
\item \textbf{Q1:} Does \modelname perform well on inductive KG reasoning?
\item \textbf{Q2:} How does each designed module influence the performance?
\item \textbf{Q3:} Whether \modelname can improve reasoning ability about distant entities or not?
\end{itemize}

\subsection{Experiments Settings}
\subsubsection{Datasets}
We perform inductive KG reasoning experiments on the benchmark datasets proposed in GraIL~\cite{grail}, which are derived from WN18RR~\cite{trans_wn18rr}, FB15k-237~\cite{trans_fb237}, and NELL-995~\cite{trans_nell_995}.
Each benchmark dataset is divided into four versions (v1, v2, v3, v4), and the size typically increases following the version number.
Each version consists of training and test graphs without overlapping entities.
The training graphs contain triplets for training and validation, following a split ratio of 10:1.
The statistics of the datasets are presented in Table~\ref{table:inductive_datasets}.

\begin{table}[!t]
\centering
\caption{Statistics of the inductive datasets.
$\mathcal{G}$ and $\mathcal{G}^{\prime}$ denote the KGs in the training and test sets, respectively.}
\label{table:inductive_datasets}
\setlength{\tabcolsep}{4pt}
\begin{tabular}{ccccccccccc}
\toprule
\multicolumn{2}{c}{Datasets} &  \multicolumn{3}{c}{FB15k-237} & \multicolumn{3}{c}{NELL-995} & \multicolumn{3}{c}{WN18RR} \\
\cmidrule(lr){1-2}\cmidrule(lr){3-5}\cmidrule(lr){6-8}\cmidrule(lr){9-11} Versions & KGs & $|\mathcal{R}|$ & $|\mathcal{V}|$ & $|\mathcal{F}|$  & $|\mathcal{R}|$ & $|\mathcal{V}|$ & $|\mathcal{F}|$ & $|\mathcal{R}|$ & $|\mathcal{V}|$ & $|\mathcal{F}|$ \\
\midrule
\multirow{2}{*}{v1} & $\mathcal{G}$ & 183 & 2,000 & \ \,5,226 & \ \,14 & 10,915 & \ \,5,540 & \ \,9 & \ \,2,746 & \ \,6,678 \\
 & $\mathcal{G}^{\prime}$ & 146 & 1,500 & \ \,2,404 & \ \,14 & \ \ \ \,225 & \ \,1,034 & \ \,9 & \ \ \ \,922 & \ \,1,991 \\
\midrule
\multirow{2}{*}{v2} & $\mathcal{G}$ & 203 & 3,000 & 12,085 & \ \,88 & \ \,2,564 & 10,109 & 10 & \ \,6,954 & 18,968 \\
 & $\mathcal{G}^{\prime}$ & 176 & 2,000 & \ \,5,092 & \ \,79 & \ \,4,937 & \ \,5,521 & 10 & \ \,2,923 & \ \,4,863 \\
\midrule
\multirow{2}{*}{v3} & $\mathcal{G}$ & 218 & 4,000 & 22,394 & 142 & \ \,4,647 & 20,117 & 11 & 12,078 & 32,150 \\
 & $\mathcal{G}^{\prime}$ & 187 & 3,000 & \ \,9,137 & 122 & \ \,4,921 & \ \,9,668 & 11 & \ \,5,084 & \ \,7,470 \\
\midrule
\multirow{2}{*}{v4} & $\mathcal{G}$ & 222 & 5,000 & 33,916 & \ \,77 & \ \,2,092 & \ \,9,289 & \ \,9 & \ \,3,861 & \ \,9,842 \\
 & $\mathcal{G}^{\prime}$ & 204 & 3,500 & 14,554 & \ \,61 & \ \,3,294 & \ \,8,520 & \ \,9 & \ \,7,208 & 15,157 \\
\bottomrule
\end{tabular}
\end{table}

\subsubsection{Baselines}
We compare \modelname with 10 inductive baselines organized into three groups, including
(i) three rule-based models: RuleN~\cite{rulen}, Neural LP~\cite{neural_lp}, and DRUM~\cite{drum};
(ii) two subgraph-based models:  GraIL~\cite{grail} and CoMPILE~\cite{compile};
(iii) five C-MPNN-based models: NBFNet~\cite{NBFNet}, A*Net~\cite{A_Net}, RED-GNN~\cite{RED_GNN}, AdaProp~\cite{AdaProp}, and RUN-GNN~\cite{RUN_GNN}.

\subsubsection{Evaluation and Tie Policy}
Following~\cite{A_Net, RED_GNN, AdaProp}, we evaluate all the models using the filtered mean reciprocal rank (MRR) and Hits@10 metrics.
The best models are chosen according to MRR on the validation dataset.
Subgraph-based models typically rank each test triplet among 50 randomly sampled negative triplets, whereas C-MPNNs evaluate each triplet against all possible candidates.
In this paper, we follow the latter and take the results of rule-based and subgraph-based models from~\cite{AdaProp}.
Missing results are reproduced by their official code.

There are different tie policies~\cite{tie_policy} to compute MRR when several candidate entities receive equal scores.
In progressive propagation, all unvisited entities are assigned identical scores. 
Following~\cite{RED_GNN, RUN_GNN}, we measure the average rank among the entities in the tie, as suggested in~\cite{tie_policy2}.
To keep the tie policy consistent, we re-evaluate AdaProp using the official code.

\subsubsection{Implementation Details}
We implement our model using the PyTorch framework~\cite{pytorch} and employ the Adam optimizer~\cite{adam} for training.
Due to the relatively small size of the inductive dataset and its susceptibility to overfitting, we apply early stopping to mitigate this issue.
We tune the hyper-parameters using grid search and select the number of starting entities $n$ in $\{1, 2, 4, 8, 16, 32, 64\}$, the number of starting entity types $m$ in $\{2, 3, 5, 7, 9\}$.
The best hyperparameters are selected according to the MRR metric on the validation sets.
All experiments are conducted on a single NVIDIA RTX A6000 GPU with 48GB memory.

\begin{table}[!t]
\centering
\caption{Inductive KG reasoning results (measured with MRR). 
The best scores are in \textbf{bold} and the second-best scores are {\ul{underlined}}. 
``-'' denotes the result unavailable, and values with suffix ``$\star$'' are reproduced using the released code.}
\label{table:main_mrr}
\setlength{\tabcolsep}{0pt}
\begin{tabular}{lcccccccccccc}
\toprule
\multirow{2}{*}{Models} & \multicolumn{4}{c}{FB15k-237} & \multicolumn{4}{c}{NELL-995} & \multicolumn{4}{c}{WN18RR} \\
\cmidrule(lr){2-5}\cmidrule(lr){6-9}\cmidrule(lr){10-13} & v1 & v2 & v3 & v4  & v1 & v2 & v3 & v4  & v1 & v2 & v3 & v4 \\
\midrule
RuleN & .363 & .433 & .439 & .429 & .615 & .385 & .381 & .333 & .668 & .645 & .368 & .624 \\
Neural LP & .325 & .389 & .400 & .396 & .610 & .361 & .367 & .261 & .649 & .635 & .361 & .628 \\
DRUM & .333 & .395 & .402 & .410 & .628 & .365 & .375 & .273 & .666 & .646 & .380 & .627 \\
\midrule
GraIL & .279 & .276 & .251 & .227 & .481 & .297 & .322 & .262 & .627 & .625 & .323 & .553 \\
CoMPILE & .287 & .276 & .262 & .213 & .330 & .248 & .319 & .229 & .577 & .578 & .308 & .548 \\
\midrule
NBFNet & .270 & .321 & .335 & .288 & .584 & .410 & .425 & .287 & .686 & .662 & .410 & .601 \\
A*Net & - & - & - & - & - & - & - & - & - & - & - & - \\
RED-GNN & .341 & .411 & .411 & .421 & \ .591$^{\star}$ & \ .373$^{\star}$ & \ .391$^{\star}$ & \ .195$^{\star}$ & .693 & .687 & .422 & .642 \\
AdaProp & \ .279$^{\star}$ & \ .467$^{\star}$ & \ {\ul .470$^{\star}$} & \ .440$^{\star}$ & \ {\ul .725$^{\star}$} & \ {\ul .416$^{\star}$} & \ .413$^{\star}$ & \ {\ul .338$^{\star}$} & \ {\ul .706$^{\star}$} & \ \textbf{.703}$^{\star}$ & \ .433$^{\star}$ & \ {\ul .651$^{\star}$} \\
RUN-GNN & {\ul .397} & {\ul .473} & .468 & {\ul .463} & \ $.617^{\star}$ & \ $.413^{\star}$ & \ {\ul $.479^{\star}$} & \ $.282^{\star}$ & .699 & .697 & \textbf{.445} & \textbf{.654} \\

\midrule
\modelname & \textbf{.458} & \textbf{.526} & \textbf{.506} & \textbf{.487} & \textbf{.787} & \textbf{.540} & \textbf{.496} & \textbf{.384} & \textbf{.733} & {\ul .702} & {\ul .442} & .645\\
\bottomrule
\end{tabular}
\end{table}

\begin{table}[!t]
\centering
\caption{Inductive KG reasoning results (measured with Hits@10)}
\label{table:main_hits@10}
\setlength{\tabcolsep}{0pt}
\begin{tabular}{lcccccccccccc}
\toprule
\multirow{2}{*}{Models} & \multicolumn{4}{c}{FB15k-237} & \multicolumn{4}{c}{NELL-995} & \multicolumn{4}{c}{WN18RR} \\
\cmidrule(lr){2-5}\cmidrule(lr){6-9}\cmidrule(lr){10-13} & v1 & v2 & v3 & v4  & v1 & v2 & v3 & v4  & v1 & v2 & v3 & v4  \\
\midrule
RuleN & .446 & .599 & .600 & .605 & .760 & .514 & .531 & .484 & .730 & .694 & .407 & .681 \\
Neural LP & .468 & .586 & .571 & .593 & .871 & .564 & .576 & .539 & .772 & .749 & .476 & .706 \\
DRUM & .474 & .595 & .571 & .593 & {\ul .873} & .540 & .577 & .531 & .777 & .747 & .477 & .702 \\
\midrule
GraIL & .429 & .424 & .424 & .389 & .565 & .496 & .518 & .506 & .760 & .776 & .409 & .687 \\
CoMPILE & .439 & .457 & .449 & .358 & .575 & .446 & .515 & .421 & .747 & .743 & .406 & .670 \\
\midrule
NBFNet & .530 & .644 & .623 & .642 & .795 & {\ul .635} & .606 & {\ul .591} & \textbf{.827} & .799 & \textbf{.568} & .702 \\
A*Net & {\ul .535} & .638 & .610 & .630 & - & - & - & - & .810 & \textbf{.803} & .544 & \textbf{.743} \\
RED-GNN & .483 & .629 & .603 & .621 &\  .866$^{\star}$ &\  .601$^{\star}$ &\  .594$^{\star}$ &\  .556$^{\star}$ & .799 & .780 & .524 & .721 \\
AdaProp & \ .461$^{\star}$ & \ {\ul .665$^{\star}$} &\ {\ul .636$^{\star}$} &\  .632$^{\star}$ &\  .776$^{\star}$ &\  .618$^{\star}$ &\  .580$^{\star}$ &\  .589$^{\star}$ &\  .796$^{\star}$ &\  .792$^{\star}$ &\ .532$^{\star}$ &\ .730$^{\star}$ \\
RUN-GNN & .496 & .639 & .631 & \textbf{.665} &\ .833$^{\star}$ &\  .575$^{\star}$ &\  {\ul .659$^{\star}$} &\  .436$^{\star}$ & .807 & .798 & {\ul .550} & {\ul .735} \\
\midrule
\modelname & \textbf{.583} & \textbf{.702} & \textbf{.675} & \textbf{.665} & \textbf{.900} & \textbf{.735} & \textbf{.666} & \textbf{.617} & {\ul .817} & \textbf{.803} & .547 & .726 \\
\bottomrule
\end{tabular}
\end{table}

\subsection{Main Results (Q1)}
Tables~\ref{table:main_mrr} and \ref{table:main_hits@10} depict the performance of different models on inductive KG reasoning.
\modelname demonstrates the best performance across all metrics on FB15k-237 and NELL-995, and compares favorably with the top models on WN18RR.
We observe that (i) subgraph-based models typically perform poorly.
This is because subgraphs are often sparse or empty and provide less information, particularly for distant entities.
(ii) Rule-based models are generally more competitive but are still weaker compared to C-MPNN-based models.
However, DRUM outperforms existing models except \modelname in Hits@10 on NELL-995~(v1).
NELL-995~(v1) is a special dataset and the distance between the head and tail entities for all triplets in the test graph is no longer than 3, which is very short.
Thus, we conjecture that the length of the learned rules limits the reasoning capabilities  of rule-based models.
Differently, \modelname holds an edge over these two groups of models on all datasets.
This suggests that multiple starting entities in \modelname alleviate the distance limit issues as much as possible when reasoning.

Compared with the best C-MPNN-based results, \modelname achieves an average relative gain of 9.9\% in MRR, 5.2\% in Hits@10 on FB15k-237, and 13.9\% in MRR, 6.1\% in Hits@10 on NELL-995.
Existing C-MPNN-based models typically use all entities in the KG or only the head entity as starting entities, without providing conditional information to distant entities, which can introduce excessive noise or lack sufficient information.
Instead, our \modelname selects multiple query-dependent starting entities adaptively and propagates conditions farther through the highway for accurate reasoning.
Moreover, LinkVerify in \modelname additionally reduces noisy samples in training.
We also observe that the improvement of the model on WN18RR is not as pronounced as on the other datasets.
To provide insights into this phenomenon, we conduct further analysis in Section~\ref{section:further_analysis}.

\subsection{Ablation Study}

\subsubsection{Variants of \modelname (Q2)}
In this section, we design several variants of \modelname to study the contributions of three components: (i) selection, (ii) highway, and (iii) LinkVerify in training.
The results are summarized in Tables~\ref{table:ablation_mrr} and \ref{table:ablation_hits@10}, which indicate that all components contribute significantly to \modelname on the three datasets.

First, the variant of w/o selection propagates only from the head entity which is the same as RED-GNN.
According to the results, removing selection significantly decreases performance, highlighting the effectiveness of using multiple starting entities to explore reasoning patterns across a broader neighborhood.

Second, it can be observed that the performance of variant w/o highway is worse than \modelname.
This observation suggests that transmitting query-dependent information to the starting entities is a promising approach to expedite propagation for conditions and enhance reasoning accuracy.

Third, the variant of w/o LinkVerify is inferior to \modelname all the time, as triplets with unvisited target entities in training KG introduce noise.
Removing LinkVerify results in poorer performance, especially on smaller datasets. 
For instance, w/o LinkVerify decreases 7.0\% for FB15k-237 (v1) and 1.3\% for FB15k-237 (v4) relatively.
This is because the noisy triplets negatively influence training when data is lacking.
Thus, LinkVerify demonstrates to be more effective when applied to KGs with fewer triplets.

\begin{table}[!t]
\centering
\caption{Ablation study of the proposed framework (measure with MRR)}
\label{table:ablation_mrr}
\setlength{\tabcolsep}{2pt}
\begin{tabular}{lcccccccccccc}
\toprule
\multirow{2}{*}{Models} & \multicolumn{4}{c}{FB15k-237} & \multicolumn{4}{c}{NELL-995} & \multicolumn{4}{c}{WN18RR} \\
\cmidrule(lr){2-5}\cmidrule(lr){6-9}\cmidrule(lr){10-13} & v1 & v2 & v3 & v4  & v1 & v2 & v3 & v4  & v1 & v2 & v3 & v4 \\
\midrule
\modelname & \textbf{.458} & \textbf{.526} & \textbf{.506} & \textbf{.487} & \textbf{.787} & \textbf{.540} & \textbf{.496} & \textbf{.384} & \textbf{.733} & \textbf{.702} & \textbf{.442} & \textbf{.645} \\
\ \ w/o Selection & .432 & .491 & .483 & .457 & .719 & .479 & .457 & .280 & .721 & .674 & .432 & .643 \\
\ \ w/o Highway & .411 & .488 & .460 & .474 & .774 & .473 & .494 & .297 & .726 & .700 & .438 & .629 \\
\ \ w/o LinkVerify & .426 & .517 & .498 & .481 & .661 & .502 & .482 & .375 & .729 & .698 & .420 & .641 \\
\bottomrule
\end{tabular}
\end{table}

\begin{table}[!t]
\centering
\caption{Ablation study of the proposed framework (measured with Hits@10)}
\label{table:ablation_hits@10}
\setlength{\tabcolsep}{2pt}
\begin{tabular}{lcccccccccccc}
\toprule
\multirow{2}{*}{Models} & \multicolumn{4}{c}{FB15k-237} & \multicolumn{4}{c}{NELL-995} & \multicolumn{4}{c}{WN18RR} \\
\cmidrule(lr){2-5}\cmidrule(lr){6-9}\cmidrule(lr){10-13} & v1 & v2 & v3 & v4 & v1 & v2 & v3 & v4 & v1 & v2 & v3 & v4 \\
\midrule
\modelname & \textbf{.583} & \textbf{.702} & \textbf{.675} & \textbf{.665} & \textbf{.900} & \textbf{.735} & \textbf{.666} & \textbf{.617} & \textbf{.817} & \textbf{.803} & \textbf{.547} & \textbf{.726} \\
\ \ w/o Selection & .534 & .686 & .644 & .629 & .775 & .693 & .619 & .425 & .811 & .778 & .528 & .717 \\
\ \ w/o Highway & .532 & .657 & .609 & .644 & .855 & .682 & .648 & .532 & .814 & .788 & .543 & .698 \\
\ \ w/o LinkVerify & .568 & .699 & .657 & .658 & .785 & .695 & .645 & .608 & .811 & .797 & .508 & .724 \\
\bottomrule
\end{tabular}
\end{table}

\begin{table}[!t]
\centering
\caption{Per-distance evaluation on FB15k-237 (v1) (measured with Hits@10). ``$\infty$'' indicates that the head entity fails to reach the tail entity.}
\label{table:ablation_distance}
\setlength{\tabcolsep}{4pt}
\begin{tabular}{ccccccc}
\toprule
Distance & Proportions & RED-GNN & AdaProp & RUN-GNN & NBFNet & \modelname \\
\midrule
\ \,1 & 32.68\% & .813 & .933 & .851 & .545 & .948 \\
\ \,2 & 12.20\% & .640 & .520 & .740 & .760 & .780 \\
\ \,3 & 25.37\% & .433 & .269 & .414 & .490 & .471 \\
\ \,4 & \ \,7.32\% & .000 & .000 & .267 & .333 & .300 \\
\ \,5 & 11.22\% & .000 & .000 & .217 & .261 & .174 \\
\ \,6 & \ \,3.90\% & .000 & .000 & .000 & .438 & .188 \\
\ \,7 & \ \,1.46\% & .000 & .000 & .000 & .333 & .000 \\
\ \,8 & \ \,1.46\% & .000 & .000 & .000 & .333 & .167 \\
\ \,9 & \ \,0.00\% & .000 & .000 & .000 & .000 & .000 \\
10 & \ \,0.98\% & .000 & .000 & .000 & .250 & .000 \\
$\infty$ & \ \,3.41\% & .000 & .000 & .000 & .357 & .214
\\
\bottomrule
\end{tabular}
\end{table}

\subsubsection{Per-distance Performance (Q3)}
To check the reasoning ability on distant tail entities, we compare \modelname with several expressive models on FB15k-237 (v1).
To make the comparison more precise, we split FB15k-237 (v1) into 11 subsets according to the shortest distance between the head and tail entity for each triplet.
The comparisons are conducted on each subset based on official code and parameters.
RED-GNN, AdaProp and \modelname use 3 layers of GNN.
RUN-GNN and NBFNet use 5 and 6 layers of GNN, respectively.
The results are shown in Table~\ref{table:ablation_distance}.

Compared to the models with a single starting entity (RED-GNN, AdaProp, and RUN-GNN), \modelname performs better significantly on distant entities.
For instance, RED-GNN fails to predict entities beyond 3 hops.
Moreover, \modelname can even reason about unreachable target entities.
This is because \modelname can select query-related starting entities that are disconnected from the head entity but in the neighborhood of the unreachable entities.
These observations demonstrate that multiple starting entities can expand the reasoning area effectively, and the highway layer provides additional evidence for reasoning about distant entities.

Differently, the reasoning performance of NBFNet on close entities is significantly decreased despite the ability to reason about distant entities.
For instance, NBFNet is inferior to the other models on Hits@10 for 1-distance triplets with a great gap of at least 0.268.
This is because NBFNet propagates from query-independent starting entities and reasons along many noisy relational paths, which disrupts the inference about close entities.
Instead, \modelname improves the reasoning performance for distant entities and keeps the reasoning abilities for close entities simultaneously.
This is achieved due to MStar propagating conditions along query-related relational paths and removing noisy links by LinkVerify.

\begin{table}[!t]
\centering
\caption{Proportions of long-distance triplets in the KGs.
The shortest distance between head and tail entities in a long-distance triplet is longer than 3.}
\label{table:three_hops_ratio}
\setlength{\tabcolsep}{4pt}
\begin{tabular}{ccccccc}
\toprule
\multicolumn{1}{c}{Datasets} & \multicolumn{2}{c}{FB15k-237} & \multicolumn{2}{c}{NELL-995} & \multicolumn{2}{c}{WN18RR} \\
\cmidrule(lr){1-1}\cmidrule(lr){2-3}\cmidrule(lr){4-5}\cmidrule(lr){6-7} Versions & $\mathcal{G}$ & $\mathcal{G}^{\prime}$ & $\mathcal{G}$ & $\mathcal{G}^{\prime}$ & $\mathcal{G}$ & $\mathcal{G}^{\prime}$ \\
\midrule
v1 & 15.78\% & 29.76\% & 39.64\% & \ \,\textbf{0.00}\% & 34.31\% & 17.55\% \\
v2 & \ \,8.69\% & 15.48\% & 10.62\% & \ \,2.52\% & 20.86\% & 16.33\% \\
v3 & \ \,3.41\% & \ \,4.51\% & 11.16\% & \ \,3.96\% & 22.32\% & 26.94\% \\
v4 & \ \,2.39\% & \ \,2.74\% & \ \,9.30\% & \ \,6.98\% & 22.39\% & 20.50\% \\
\bottomrule
\end{tabular}
\end{table}

\subsection{Further Analysis}
\label{section:further_analysis}
\subsubsection{Perspective of Datasets}
As shown in Tables~\ref{table:ablation_mrr} and~\ref{table:ablation_hits@10}, the improvement of \modelname on WN18RR is not as great as the one on other datasets.
As can be seen from Table~\ref{table:inductive_datasets}, WN18RR~(v1, v2, v3, v4) and NELL-995~(v1) have fewer relations.
Due to the entity-independent nature of inductive KG reasoning, entity embeddings usually rely on the representation of relations.
With fewer relations, entities carry more monotonous information.
Therefore, it becomes challenging to select query-dependent entities and propagate messages to the target ones. 
To study the situation further, we count the proportion of triplets whose shortest distance between the head and tail entities exceeds 3.
We regard these triplets as long-distance triplets.
The result is shown in Table~\ref{table:three_hops_ratio}.
We can see that NELL-995~(v1) owns zero long-distance triplets in the test graphs.
Thus, NELL-995~(v1) can resolve the above issues by propagating conditional information to any target entity in 3 hops, even without multiple starting entities.

\subsubsection{Perspective of Starting Entities Selection}
\modelname leverages an importance score function to select starting entities.
The score function is conditioned on the query head and relation, aiming to explore query-dependent entities.
Here, we consider two other score function variants, i.e., variant w/~random and variant w/~degree.
Variant w/~random scores the entities with random values.
Similar to EL-GNN~\cite{el_gnn}, variant w/ degree assigns higher scores to entities with higher degrees.
All variants keep top-$n$ entities as starting ones.

\begin{table}[!t]
\centering
\caption{Comparison of different starting entities selection methods}
\label{table:variants_root}
\setlength{\tabcolsep}{4pt}
\begin{tabular}{lcccccc}
\toprule
\multirow{2}{*}{Models} & \multicolumn{2}{c}{FB15k-237 (v1)} & \multicolumn{2}{c}{NELL-995 (v1)} & \multicolumn{2}{c}{WN18RR (v1)} \\
\cmidrule(lr){2-3}\cmidrule(lr){4-5}\cmidrule(lr){6-7} & MRR & Hits@10 & \multicolumn{1}{c}{MRR} & \multicolumn{1}{c}{Hits@10} & \multicolumn{1}{c}{MRR} & \multicolumn{1}{c}{Hits@10} \\
\midrule
\modelname & \textbf{.462} & \textbf{.598} & \textbf{.801} & \textbf{.921} & \textbf{.736} & \textbf{.816} \\
\ \ w/ random & .427 & .587 & .787 & .901 & .698 & .803\\
\ \ w/ degree & .403 & .553 & .362 & .595 & .709 & .810
\\
\bottomrule
\end{tabular}
\end{table}

Table~\ref{table:variants_root} shows the comparison results.
We can observe that random scores lead to a degraded performance.
This is because random starting entities propagate along many noisy relational paths.
Noisy paths hinder \modelname's ability to capture query-related rules and to reach distant target tail entities.
Variant w/~degree is also inferior to our \modelname, even worse than random scores.
For instance, the performance of variant w/~degree on FB15k-237~(v1) decreases by 54.8\% and 54.0\% relative to \modelname and variant w/ random, respectively.
This is mainly due to the fact that the global feature degree fixes the starting entities and cannot support query-dependent propagation.

\section{Conclusion and Future Work}
\label{section:conclusion}
In this paper, we explore the issue of inefficient message propagation for KG reasoning and propose a new inductive KG reasoning model called \modelname.
Specifically, we propose using multiple starting entities to expand the propagation scope.
Moreover, we construct a highway between the head entity and the other starting entities to accelerate conditional message passing.
Additionally, we introduce a training strategy LinkVerify to filter inappropriate samples.
Experimental results demonstrate the effectiveness of \modelname.
In particular, ablation results validate the superiority of \modelname for reasoning about distant entities.
In future work, we plan to explore alternative modules for selecting and classifying starting entities. 
We also intend to investigate methods to effectively utilize noisy triplets during training instead of dropping them.

\subsubsection*{Acknowledgments}
We thank the anonymous reviewers for their valuable comments.
This work was supported by the National Natural Science Foundation of China under Grant 62272219 and the Collaborative Innovation Center of Novel Software Technology \& Industrialization.

\subsubsection*{Supplemental Material Statement}
The source code and hyperparameters are available at our GitHub repository: 
\url{https://github.com/nju-websoft/MStar}.

\bibliographystyle{refs/splncs04} 
\bibliography{refs/reference}
\end{document}